\documentclass[letterpaper, 10 pt, journal, twoside]{IEEEtran}

\usepackage{relsize}
\usepackage{xspace}
\usepackage{color}
\usepackage{graphics}
\usepackage{graphicx}

\usepackage{amsmath}

\usepackage{amsfonts}
\usepackage{amssymb}
\usepackage{amsthm}

\usepackage{float}
\usepackage{url}
\usepackage{booktabs}
\usepackage{multirow}
\usepackage{hhline}
\usepackage[capitalise]{cleveref}
\usepackage{dsfont}
\usepackage{cancel}

\usepackage{subcaption}
\usepackage{catchfilebetweentags}

\usepackage{booktabs}

\usepackage{csvsimple}

\usepackage{tikz}
\usetikzlibrary{positioning}

\usepackage{pgfplots}
\pgfplotsset{compat=1.18}  

\usepackage{pifont}

\usepackage{bm}

\newcommand{\bdmath}{\begin{dmath}}
\newcommand{\edmath}{\end{dmath}}
\newcommand{\beq}{\begin{equation}}
\newcommand{\eeq}{\end{equation}}
\newcommand{\bdm}{\begin{displaymath}}
\newcommand{\edm}{\end{displaymath}}
\newcommand{\bea}{\begin{eqnarray}}
\newcommand{\eea}{\end{eqnarray}}
\newcommand{\beal}{\beq \begin{array}{ll}}
\newcommand{\eeal}{\end{array} \eeq}
\newcommand{\beas}{\begin{eqnarray*}}
\newcommand{\eeas}{\end{eqnarray*}}
\newcommand{\ba}{\begin{array}}
\newcommand{\ea}{\end{array}}
\newcommand{\bit}{\begin{itemize}}
\newcommand{\eit}{\end{itemize}}
\newcommand{\ben}{\begin{enumerate}}
\newcommand{\een}{\end{enumerate}}

\newcommand{\M}[1]{{\bm #1}} 

\newcommand{\hide}[1]{}

\newcommand{\hiddenText}{{\color{gray} hidden text.}}
\newcommand{\hideWithText}[1]{\hiddenText}

\newcommand{\SEthree}{\ensuremath{\mathrm{SE}(3)}\xspace}

\newcommand{\SOthree}{\ensuremath{\mathrm{SO}(3)}\xspace}

\newcommand{\blue}[1]{{\color{blue}#1}}

\newcommand{\linkToPdf}[1]{\href{#1}{\blue{(pdf)}}}
\newcommand{\linkToPpt}[1]{\href{#1}{\blue{(ppt)}}}
\newcommand{\linkToCode}[1]{\href{#1}{\blue{(code)}}}
\newcommand{\linkToWeb}[1]{\href{#1}{\blue{(web)}}}
\newcommand{\linkToVideo}[1]{\href{#1}{\blue{(video)}}}
\newcommand{\award}[1]{\xspace} 

\usepackage[acronym]{glossaries}
\newacronym{cslam}{C-SLAM}{Collaborative Simultaneous Localization and Mapping}
\newacronym{slam}{SLAM}{Simultaneous Localization and Mapping}
\newacronym{mold}{FLOCC-SLAM}{Foundation model-based LOop Closing Collaborative SLAM}

\begin{document}

\title{3D Foundation Model-Based Loop Closing for Decentralized Collaborative SLAM}

\author{Pierre-Yves Lajoie$^{1}$, Benjamin Ramtoula$^{2}$, Daniele De Martini$^{2*}$, Giovanni Beltrame$^{1*}$
\thanks{  This work was partially supported by a Vanier Canada Graduate Scholarships
   Award, a MITACS Globalink Award, the EPSRC Centre for Doctoral Training in Autonomous Intelligent Machines and Systems [EP/S024050/1], and Oxa. \;\;\;\; $^{*}$Equal supervision. $^{1}$P.\,Lajoie and G.\,Beltrame are with the Department of Computer and Software Engineering, \mbox{Polytechnique Montr\'eal}, Montreal, Canada,
 {\tt\scriptsize\,\{pierre-yves.lajoie, giovanni.beltrame\}@polymtl.ca}.}
 \thanks{$^{2}$B.\,Ramtoula and D.\,De Martini are with the Oxford Robotics Institute, Dept. of Engineering Science, University of Oxford, UK,
 {\tt\scriptsize\,\{benjamin, daniele\}@robots.ox.ac.uk}.}
 }

\maketitle

\begin{abstract}
    Decentralized \gls{cslam} techniques often struggle to identify map overlaps due to significant viewpoint variations among robots.
Motivated by recent advancements in 3D foundation models, which can register images despite large viewpoint differences, we propose a robust loop closing approach that leverages these models to establish inter-robot measurements.
In contrast to resource-intensive methods requiring full 3D reconstruction within a centralized map, our approach integrates foundation models into existing SLAM pipelines, yielding scalable and robust multi-robot mapping.
Our contributions include: (1) integrating 3D foundation models to reliably estimate relative poses from monocular image pairs within decentralized \gls{cslam}; (2) introducing robust outlier mitigation techniques critical to the use of these relative poses; and (3) developing specialized pose graph optimization formulations that efficiently resolve scale ambiguities. We evaluate our method against state-of-the-art approaches, demonstrating improvements in localization and mapping accuracy, alongside significant gains in computational and memory efficiency.
These results highlight the potential of our approach for deployment in large-scale multi-robot scenarios.
\end{abstract}
\glsresetall


\section{Introduction}
\label{sec:introduction}

\IEEEPARstart{D}{ecentralized} \gls{cslam} is a critical capability for multi-robot systems
operating in unknown environments. In these scenarios, multiple robots must explore the
environment independently, while exchanging information to build a shared global
map. This task becomes particularly challenging when the robots' viewpoints
differ significantly due to varying trajectories and diverse sensor placements,
making it difficult to indentify overlapping map sections and generate reliable inter-robot
loop closures~\cite{lowryVisualPlaceRecognition2016}.
Moreover, when network issues or bandwidth limitations prevent centralized processing, loop closure detection must occur between the agents with minimal data transmission, as exchanging entire maps is too bandwidth-intensive.
To address these challenges, we introduce a novel loop closing approach that leverages recent advancements in 3D foundation models~\cite{wangDUSt3RGeometric3D2024,leroyGroundingImageMatching2024} to 
 handle extreme viewpoint variations without map sharing to find overlaps.
Specifically, our approach capitalizes on the ability of 3D foundation models to perform up-to-scale relative pose estimation from pairs of monocular images, even in cases of opposite viewpoints and unknown image domains thanks to extensive pretraining on large-scale datasets.
We name our solution \gls{mold}.

\gls{mold} is designed for seamless integration with existing single-robot odometry systems, enabling robust, efficient, and easily deployable C-SLAM on top of individual robot localization pipelines, assuming these provide locally consistent, metric-scale odometry (e.g., VIO).
During pose graph optimization, our method utilizes the metric scale from odometry to infer the correct scale of inter-robot measurements obtained from 3D foundation models.

This paper introduces three main contributions:
\begin{itemize}
\item the integration of a 3D foundation model
  (MASt3R~\cite{leroyGroundingImageMatching2024}) into a decentralized \gls{cslam} pipeline
  to estimate relative poses between robots based on monocular image pairs,
  providing a means to detect inter-robot loop closures in situations with
  limited viewpoint overlap;
\item outlier detection and uncertainty modelling designed for 3D
  relative pose estimation with MASt3R, to reduce the occurrence of spurious loop closures;
\item a set of specialized pose-graph optimization formulations to merge
  individual robot maps, resolving 3D scale ambiguities of the generated loop
  closures and refining the overall localization accuracy.
\end{itemize}

Overall, our contributions enable us to leverage recent advances in 3D
foundation models to enhance the robustness of \gls{cslam} and its performance
across a wider range of environments and multi-robot missions. We evaluate the
performance of our approach against state-of-the-art decentralized \gls{cslam}
algorithms on several multi-robot dataset sequences.
Our results demonstrate
that the powerful representations generated by 3D foundation models enable
substantial improvements in localization accuracy while reducing computational
and memory overhead through specialized optimization and keyframe
sparsification.
This makes our approach a promising solution for large-scale
multi-robot deployments in unknown environments, where inter-robot collaboration
is crucial for efficient exploration and mapping.



\section{Background and Related Works}
\label{sec:related_work}

In \gls{cslam}, multiple robots work together to build a shared map and localize
within it. By sharing sensor data and detecting overlap between their maps, the
robots can improve their individual localization and create a globally
consistent view of the environment across the robots. In this section, we
present the background and related work on \gls{cslam}, as well as on
image-based relative-pose estimation.

\subsubsection{Collaborative SLAM}

\glsunset{slam} In \gls{cslam}, robots typically perform \gls{slam} individually
and then share information about their maps to fuse them into a globally
consistent estimate of the traversed environment. Similar to single-robot
\gls{slam}, \gls{cslam} is typically divided in two parts: the front-end, which
is responsible for feature extraction and data association, and the back-end,
which manages state estimation~\cite{cadenaPresentFutureSimultaneous2016}.

One of the most challenging task of the front-end is to efficiently detect and
compute inter-robot loop closures within the overlapping regions.
These loop closures correspond to
connections between independent robots' estimates that can be discovered when
the same places are visited by different robots. They serve as stitching points
to merge local maps into a global representation of the environment. To
efficiently merge large maps, loop closure detection is typically performed in
two stages: place recognition, to
identify possible map overlaps, followed by the registration to compute the
3D relative pose between the individual overlaps.

The back-end then estimates the most likely poses and map based on
measurements collected from all robots. Our work focuses on pose-graph
formulations of \gls{slam}, where features are marginalized into inter-pose
measurements, as this approach is generally more efficient for larger
maps~\cite{cadenaPresentFutureSimultaneous2016}.

However, the perceptual aliasing phenomenon, where distinct places are mistaken
for the same location, can cause front-end techniques to produce
spurious measurements. Several methods have been proposed to address this pervasive issue in \gls{slam},
such as Pairwise Consistency
Maximization~\cite{mangelsonPairwiseConsistentMeasurement2018}, or Graduated Non-Convexity
(GNC)~\cite{yangGraduatedNonConvexityRobust2020b}.

In this work, we specifically focus on key challenges associated with decentralized \gls{cslam}, specifically generating pairwise inter-robot loop closures and maintaining scale consistency, all without exchanging complete maps.
A number of decentralized \gls{cslam} systems have been developed in recent years. DSLAM~\cite{cieslewskiDataEfficientDecentralizedVisual2018} was an early approach that leveraged compact learned descriptors to enable efficient distributed place recognition in the front-end. Subsequent systems, such as Kimera-Multi~\cite{tianKimeraMultiRobustDistributed2022}, extended this work by incorporating robust distributed back-end solvers. Another system, Swarm-SLAM~\cite{lajoieSwarmSLAMSparseDecentralized2024} built upon these advances by introducing a sparsification strategy that prioritizes inter-robot loop closures, significantly improving front-end efficiency.
Most recently, DVM~\cite{birdDVMSLAMDecentralizedVisual2025}, released concurrently with this paper, is the first system for decentralized monocular C-SLAM that executes both its front-end and back-end entirely onboard each agent.

\subsection{Relative Pose Estimation}
Producing globally consistent maps in \gls{cslam} heavily relies on the ability
to accurately perform relative pose estimation, which involves performing 3D
registration between two keyframes from different robots.

The geometric methods usually
employed in monocular
setups~\cite{hartleyMultipleViewGeometry2003} can only estimate
relative poses up-to-scale, meaning the absolute metric scale of the
transformation remains unknown. While scale information for consecutive images
can be recovered by combining image-based estimates with motion sensors such as
IMUs~\cite{qinVINSMonoRobustVersatile2018}, these sensors cannot be used to
recover the scale of relative poses between non-consecutive images.
Moreover,
compared to consecutive images, non-consecutive ones typically exhibit larger
viewpoint differences, where traditional keypoint-based matching methods—such as
those relying on hand-crafted
features—often struggle.
Recent
advancements in monocular depth estimation -- such as
DPT~\cite{ranftlVisionTransformersDense2021} -- offer promising alternatives for
scale estimation. However, despite their potential, these methods often require
domain-specific fine-tuning, which can limit their generalizability when mapping
unknown environments.

To overcome viewpoint
limitations, learning-based methods like
SuperPoint~\cite{detoneSuperPointSelfSupervisedInterest2018} and
SuperGlue~\cite{sarlinSuperGlueLearningFeature2020} use deep-learning
techniques to enhance keypoint-matching robustness, incorporating reasoning
across the entire image to improve performance. The field has also introduced
new datasets and benchmarks, such as the Map-Free
challenge~\cite{arnoldMapFreeVisualRelocalization2022} which focuses on scenarios with
drastic viewpoints and illumination changes, requiring the emergence of new
techniques to succeed, such as MicKey~\cite{barroso-lagunaMatching2DImages2024}
and DUSt3R~\cite{wangDUSt3RGeometric3D2024}. The first predicts metric
correspondences directly in 3D camera space instead of the usual 2D pixel
space, while DUSt3R reformulates the image matching problem as a pairwise 3D
reconstruction task, predicting and aligning 3D pointmaps to estimate relative
poses.
Extending DUSt3R, MASt3R~\cite{leroyGroundingImageMatching2024}
regresses local features and explicitly trains for
pairwise matching.

Building on these recent advances, MAST3R-SfM~\cite{duisterhofMASt3RSfMFullyIntegratedSolution2024} and MAST3R-SLAM~\cite{muraiMASt3RSLAMRealTimeDense2024} integrate these 3D foundation models into complete mapping and localization frameworks. Similar to traditional SfM and SLAM systems, these methods reduce computational complexity by selecting keyframes instead of processing all images, constructing a graph where keyframes with overlapping viewpoints are linked—analogous to the pose graphs commonly used in SLAM.
However, while highly effective in the single-robot case, these approaches are less effective for multi-robot SLAM because they rely on monolithic global optimization, jointly refining poses and scene geometry.
This involves centralized processing, where all keyframe images must be collected and processed on a central server, which requires continuous high-bandwidth communication with all robots.
Moreover, in multi-robot settings, the number of keyframes and constraints, from odometry and inter-robot loop closures, grows faster than with a single robot, making real-time onboard computation with large models impractical.

In contrast, our approach is specifically designed for the constraints of decentralized systems. By leveraging 3D foundation models for robust inter-robot loop closures from just a pair of images, we avoid large map exchanges between robots. Rather than relying on a monolithic global optimization, our approach distributes the computationally expensive encoding step and delegates the optimization task to a dynamically elected robot, following the strategy in~\cite{lajoieSwarmSLAMSparseDecentralized2024}, thereby mitigating the primary communication and computation bottlenecks inherent in decentralized operation.

\section{Method}
\label{sec:methodology}




Our pose-graph visual \gls{cslam} approach assumes that each robot has access to a pre-existing localization source with correct metric scale—such as those obtained from visual-inertial~\cite{qinVINSMonoRobustVersatile2018}, stereo~\cite{labbeRTABMapOpensourceLidar2019}, or LiDAR-based~\cite{shanLIOSAMTightlycoupledLidar2020} systems.
We define the input pose estimates of a robot $\alpha$ as $T_{\alpha,0}, \dots, T_{\alpha,n} \in \SEthree\mathrm{,}$
%
%
where $T_{\alpha,i}$ represents the pose estimate of robot $\alpha$ at each of the $i \in n$ keyframes along its trajectory.
Alongside each keyframe, our approach also takes as input the corresponding image frame $I_i^\alpha$, which will be used for inter-robot loop-closure detection.

\subsection{Place Recognition}
\label{subsec:placerecognition}

The first step of our pipeline is to perform place recognition between the maps of different robots.
By identifying locations visited by two or more robots, we can create inter-robot loop closures that link the individual robot pose graphs into a consistent, shared localization estimate.
For place recognition, while~\cite{duisterhofMASt3RSfMFullyIntegratedSolution2024,muraiMASt3RSLAMRealTimeDense2024} utilize MASt3R-encoded features with Aggregated Selective Match Kernels~\cite{toliasLearningAggregatingDeep2020}, we chose CosPlace~\cite{bertonRethinkingVisualGeoLocalization2022}
for its strong out-of-the-box performance. It enables us to demonstrate state-of-the-art \gls{cslam} using pretrained models without any calibration needed to adapt to a new domain, a key advantage for deployment in uncontrolled environments.


When two or more robots are within communication range, the robots exchange their compact CosPlace descriptors with one another.
Once a robot $\alpha$ has received descriptors from a neighboring robot $\beta$, it compares them with its own descriptors using cosine similarity.
For each pair of keyframes $(I^{\alpha}_{i}, I^{\beta}_{j})$, we compute the cosine similarity score $s^{\alpha, i}_{\beta, j}$.
%
The best match for each keyframe is determined through a nearest neighbor search and, if the best match similarity exceeds a threshold, the corresponding pair of keyframes $(I^{\alpha}_{i}, I^{\beta}_{j})$ is considered an inter-robot loop-closure candidate.
These candidates are then passed to the subsequent registration step, where the relative pose $T^{\alpha, i}_{\beta, j}$ between the two keyframes is computed.

\subsection{Registration}
\label{subsec:registration}

While previous techniques~\cite{tianKimeraMultiRobustDistributed2022, lajoieSwarmSLAMSparseDecentralized2024} relied on 3D registration using hand-crafted image features that are highly sensitive to viewpoint changes, we leverage recent foundation models for 3D image matching~\cite{wangDUSt3RGeometric3D2024,leroyGroundingImageMatching2024}. These models can infer relative poses between images from significantly different viewpoints, enabling the detection of more inter-robot loop closures.

\ExecuteMetaData[figures/explanation_figures.tex]{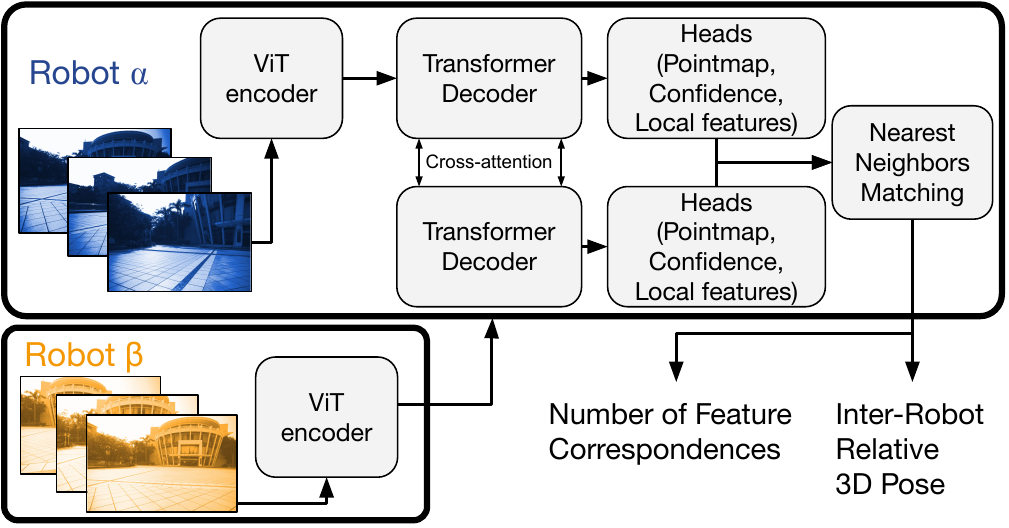}

Whereas traditional C-SLAM methods use stereo, RGB-D, or LiDAR data for relative pose estimation with accurate scale, we rely solely on monocular images and resolve the scale ambiguity at a later stage (see~\cref{subsec:pgo}). For registration, each image of a loop closure candidate $(I^{\alpha}_{i}, I^{\beta}_{j})$ is encoded locally on its corresponding robot into a compressed latent vector, using the pre-trained and frozen ViT encoder of the MASt3R model~\cite{leroyGroundingImageMatching2024}, that can later be decoded to reconstruct the scene and estimate the relative pose, as illustrated in~\cref{fig:mast3r}.
The encoding of $I^{\beta}_{j}$ is shared with robot $\alpha$ for decoding, feature matching, and inter-robot relative pose inference.
This process yields a relative pose measurement $T^{\alpha, i}_{\beta, j}$ along with the number of feature correspondences between the two frames.

As MASt3R is prone to occasional failures~\cite{leroyGroundingImageMatching2024}, we introduce a novel confidence estimation metric.
To compute it, robot $\alpha$ also performs MASt3R inference on a consecutive pair of its own images, $(I_{i-1}^{\alpha},I_{i}^{\alpha})$, for which a reliable odometry estimate exists. We then compute the ratio $r_{\beta,j}^{\alpha,i}$ of feature correspondences between the inter-robot loop closure pair $(I_{i}^{\alpha},I_{j}^{\beta})$ and the intra-robot odometry pair $(I_{i-1}^{\alpha},I_{i}^{\alpha})$:
\begin{align}
    r^{\alpha,i}_{\beta,j} = \frac{|(I^{\alpha}_{i}, I^{\beta}_{j})\;\text{feature correspondences}|}{|(I^{\alpha}_{i-i}, I^{\alpha}_{i})\;\text{feature correspondences}|}
\end{align}
Since an odometry pair typically has high overlap and many correspondences, this ratio $r_{\beta,j}^{\alpha,i}$ effectively normalizes the confidence of a loop closure match.

We use this ratio first to filter out failed registrations that fall below a threshold $R_{thr}$, and second to weight successful measurements in the pose graph. For successful loop closures, we map the ratio to a probability $p_{\beta,j}^{\alpha,i}$ using a logistic function:
\begin{align}
    p^{\alpha,i}_{\beta,j} = \left[ 1 + \text{exp}(-k \cdot (r^{\alpha,i}_{\beta,j}-1)) \right]^{-1}.
    \label{eq:confidence}
\end{align}
The probability and its parameter $k$ can be manually tuned, or learned if training data from the deployment or a similar environment is available a priori~\cite{tianResourceawareApproachCollaborative2020}.

The confidence probability is then multiplied by the information matrix associated with each measurement. The information matrix acts as a weight during pose graph optimization, ensuring that high-confidence loop closures have a greater influence on the final solution than low-confidence ones. For each inter-robot loop closure, we define the cost function $\phi_{\beta,j}^{\alpha,i}$:
\begin{align}
    \phi^{\alpha,i}_{\beta,j} &= \left\| T^{-1}_{\alpha, i} \cdot T_{\beta, j} - \bar{T}^{\alpha, i}_{\beta, j} \right\|^2_{\Omega^{\alpha, i}_{\beta, j}(p^{\alpha, i}_{\beta, j})}
\end{align}
where $\overline{T}_{\beta,j}^{\alpha,i}$ is the relative pose measurement, and $\Omega_{\beta,j}^{\alpha,i}$ is the measurement information matrix which is proportional to the confidence $p_{\beta,j}^{\alpha,i}$. This cost function can then be incorporated into a global nonlinear least-squares problem alongside odometry constraints for multi-robot pose graph optimization.


\subsection{Multi-Robot Pose Graph And Loop Scale Optimization}
\label{subsec:pgo}

In our framework, multi-robot pose graph optimization ensures that the maps and estimated trajectories of all robots are aligned within a common reference frame, providing a unified representation of the environment.
The optimization problems are formulated as a factor graph, where the variables represent the unknown quantities (e.g., the robot poses), and the factors define functions over subsets of these variables.

\paragraph{Base Multi-Robot Factor Graph}
The base optimization problem consists primarily of two types of factors: odometry factors, which link consecutive keyframe poses, and loop closure factors, which link non-consecutive keyframes.
Odometry cost functions for robot~$\alpha$ are defined as follows:
\begin{align}
    \phi^{\alpha,i-1}_{\alpha,i} &= \left\| T^{-1}_{\alpha, i-1} \cdot T_{\alpha, i} - \bar{T}^{\alpha, i-1}_{\alpha, i} \right\|^2_{\Omega_{\text{odom}}}
\end{align}
where $\bar{T}^{\alpha, i-1}_{\alpha, i}$ is the relative pose measurement, and $\Omega_{\text{odom}}$ is the corresponding information matrix.
Thus, for the multi-robot optimization problem, we minimize the sum of all cost functions:
\begin{align}
    \M{T}^{*}
    = &\operatorname*{argmin}_{\M{T}} \sum_{\alpha \in \cal{R}}  \sum_{i \in (1:n_{\alpha})} \!\!\phi^{\alpha,i-1}_{\alpha,i} + \sum_{(\alpha, i), (\beta, j) \in L_{\cal{R},\cal{R'}}} \!\!\!\!\!\!\phi^{\alpha,i}_{\beta,j}
    \label{eq:base-fg}
\end{align}
where $\cal{R}$ denotes the set of all robots, $n_{\alpha}$ represents the number of keyframes for robot $\alpha$, $L_{\cal{R},\cal{R'}}$ is the set of loop closures linking different robots, and $\M{T}$ is the set of all poses forming the robots trajectories.
The set of robots $\cal{R}$ in \cref{eq:base-fg} can involve more than two robots; in fact, we perform the optimization with all neighboring robots, detecting and incorporating inter-robot loop closures into the optimization problem for each pair.
Also, we do not consider intra-robot loop closures $\phi^{\alpha,i}_{\alpha,k}$, as we aim to isolate the effects of inter-robot loop closures, but they could straightforwardly be integrated into the formulation as an additional cost function.

The challenge with the relative poses measured with MASt3R is that their scaling is often inconsistent or very imprecise~\cite{duisterhofMASt3RSfMFullyIntegratedSolution2024,muraiMASt3RSLAMRealTimeDense2024}. Thus, to properly integrate these measurements into the optimization problem, we must either determine the scale in advance using other sensors or, as we propose here, treat the loop closure scale as an optimization variable. %
Na\"ively, we can leverage the correctly scaled odometry measurements between the two odometry poses to estimate the loop closure scale.
Assuming that the same scaling factor applies to both odometry and loop closure:
\begin{align}
    \bar{t}^{\alpha,i}_{\beta,j} = \frac{\left\|\bar{t}^{\alpha,i-1}_{\alpha,i}\right\|}{\left\|t^{\alpha,i-1}_{\alpha,i}\right\|} \cdot t^{\alpha,i}_{\beta,j}
    \label{eq:scaleinit}
\end{align}
where $t^{\alpha,i-1}_{\alpha,i}$ and $t^{\alpha,i}_{\beta,j}$ are the relative translations output by MASt3R, and $\bar{t}^{\alpha,i-1}_{\alpha,i}$ is the known translation from odometry.
While we cannot guarantee that the loop closure and odometry share the same scaling factors, this often provides a reasonable initial guess, as discussed in~\cref{subsec:ablation-scale-pgo}.

\paragraph{Independent Scales Multi-Robot Factor Graph}
In this paper, we go further and propose to explicitly optimize the loop closures' scale. For this, we draw inspiration from~\cite{lajoiePEOPLExPEdestrianOpportunistic2024}, which introduced scaling factors for pedestrian trajectories estimated via IMU dead-reckoning and refined using sporadic UWB distance measurements. While this work focused on rescaling odometry estimates, we instead apply scaling to the non-consecutive relative pose measurements.

As a first step, we decompose the measured relative pose $\bar{T}^{\alpha,i}_{\beta,j} \in \SEthree$, from MASt3R, associated with a loop closure as $\bar{R}^{\alpha,i}_{\beta,j} \in \SOthree$, and $\bar{t}^{\alpha,i}_{\beta,j} \in {\mathbb R^{3}}$, $s^{\alpha,i}_{\beta,j}$.

where $\bar{R}^{\alpha,i}_{\beta,j}$ is the measured relative rotation matrix, $\bar{t}^{\alpha,i}_{\beta,j}$ is the measured translation vector between the two poses, and $s^{\alpha,i}_{\beta,j}$ adjusts the magnitude of the translation vector to the correct scale. Together, the scaled relative pose $\hat{T}^{\alpha,i}_{\beta,j}$ is:
\begin{align}
    \hat{T}^{\alpha,i}_{\beta,j} &= \begin{bmatrix}
    \bar{R}^{\alpha,i}_{\beta,j} & s^{\alpha,i}_{\beta,j} \cdot \bar{t}^{\alpha,i}_{\beta,j}\\
    0 & 1
    \end{bmatrix}
\end{align}
We then define a new loop closure cost function that incorporates the scale value:
\begin{align}
    \hat{\phi}^{\alpha,i}_{\beta,j} &= \left\| T^{-1}_{\alpha, i} \cdot T_{\beta, j} - \hat{T}^{\alpha, i}_{\beta, j} \right\|^2_{\Omega^{\alpha, i}_{\beta, j}(p^{\alpha, i}_{\beta, j})}
\end{align}
To optimize this novel factor, we need to provide the corresponding derivatives. Specifically, since we use GTSAM~\cite{f.dellaertetal.GeorgiaTechSmoothing} as our factor graph optimization framework, we provide the analytical measurement Jacobian matrices for efficient computation, which are evaluated in the tangent space at the current estimate during optimization:
$\M{H}_{T_{\alpha, i}} = - \text{Adj}(\text{inv}({T^{-1}_{\alpha, i} \cdot T_{\beta, j}}))$,
$\M{H}_{T_{\beta, j}} = \M{I}$,
and $\M{H}_{s} = [ \M{0} -\bar{t}^{\alpha,i}_{\beta,j} ]^\top$.
Using $\hat{\phi}^{\alpha,i}_{\beta,j}$ instead of $\phi^{\alpha,i}_{\beta,j}$ in~\cref{eq:base-fg} we derive the optimization problem where the scale of each loop closure is treated independently.

\paragraph{Smoothed Scales Multi-Robot Factor Graph}
We also introduce a third formulation, which includes an additional factor that links scale factors from related loop closures to ensure they remain similar. We cluster the loop closures that link relative poses fewer than ten keyframes apart. These clusters typically occur when the robot revisits a specific area for an extended period, resulting in multiple loop closures being detected within a single large overlap between the maps. Previous work have explored identifying such clusters to detect outliers among loop closure measurements~\cite{wuClusterbasedPenaltyScaling2020}. In our case, the intuition behind linking them is that, since MASt3R is data-driven, relative poses inferred from a similar image domain might exhibit similar scaling factors.

The cost function to link the scale values, along with its corresponding Jacobians, are defined as $\phi_{s_{i,j}} = \left\|s_j - s_i\right\|^2_{\Omega_{s}}$, $\M{H}_{s_i} = -1$, and $\M{H}_{s_j} = 1$.
where the scale residual $\phi_{s_{i,j}}$ is defined as the difference between scale estimates $s_i$ and $s_j$ from two different loop closures, ensuring consistent scale across the loop closure cluster. The corresponding Jacobians, $\M{H}_{s_i}$ and $\M{H}_{s_j}$, are the partial derivatives of $\phi_{s_{i,j}}$ with respect to $s_i$ and $s_j$, respectively.
For completeness, in our experimental analysis we also consider the scenario where all loop closures share the same scale value. 
Our formulation is designed for decentralized execution. Following~\cite{lajoieSwarmSLAMSparseDecentralized2024}, optimization is carried out by a dynamically elected robot whenever a group of agents is within communication range, with the resulting estimates then broadcast to the neighboring agents.
Fully distributed solvers for our scale-aware formulations are a promising direction for future work.

\ExecuteMetaData[figures/experimentation_figures.tex]{viz-s3e}
\section{Experiments}
\label{sec:experiments}

We conducted our experiments on dataset sequences using a robot onboard computer NVIDIA Jetson AGX Xavier (32GB). We benchmark \gls{mold} against the open-source Swarm-SLAM~\cite{lajoieSwarmSLAMSparseDecentralized2024} in two configurations: stereo and LiDAR.
In the stereo configuration, place recognition was performed using CosPlace~\cite{bertonRethinkingVisualGeoLocalization2022}, and pose registration was carried out with PnP and RANSAC~\cite{labbeRTABMapOpensourceLidar2019}.
In the LiDAR configuration, place recognition utilized ScanContext~\cite{kimScanContextEgocentric2018}, and pose registration was done using TEASER++~\cite{yangTEASERFastCertifiable2021a}.
Swarm-SLAM was chosen as a strong baseline because it leverages 3D sensing, enabling us to show that our monocular-only loop-closing method can achieve comparable accuracy without relying on 3D input data.

For a thorough and fair evaluation, we also configured Swarm-SLAM with several back-end optimizers: the fast Levenberg-Marquardt solver (LM)~\cite{f.dellaertetal.GeorgiaTechSmoothing}, the robust GNC optimizer~\cite{yangGraduatedNonConvexityRobust2020b}, which can detect and reject outliers among the loop closures, and Riemannian Block Coordinate Descent (RBCD)~\cite{tianKimeraMultiRobustDistributed2022}, a distributed solver that incorporates GNC for outlier rejection.
Notably, Swarm-SLAM with RBCD in stereo mode closely resembles Kimera-Multi~\cite{tianKimeraMultiRobustDistributed2022}, except for place recognition—Kimera-Multi uses DBoW2~\cite{galvez-lopezBagsBinaryWords2012}, while Swarm-SLAM employs CosPlace~\cite{bertonRethinkingVisualGeoLocalization2022}.

For \gls{mold}, we tested three different factor graph formulations: base multi-robot pose graph, with independent scale values (\textbf{IS}), and with smoothed scale values (\textbf{SS}). Unless otherwise specified, the base formulation was used. Since the base formulation does not optimize scale, it relies only on the scale initialization technique presented in~\cref{eq:scaleinit}.
For place recognition, we set the similarity threshold $s^{\alpha, i}_{\beta, j}$ to a permissive value of $0.1$, as our pipeline robustly handles false positives in the subsequent registration stage.

We first evaluated the techniques on the S3E dataset~\cite{fengS3EMultiRobotMultimodal2024}, which features three robots navigating large, dynamic indoor and outdoor environments.
We selected the most challenging complete sequences with minimal overlap and differing viewpoints.
Additionally, we tested on the GrAco dataset~\cite{zhuGRACOMultimodalDataset2023}, which involves six robots in a large outdoor environment.
These datasets were chosen for their challenging trajectories and compatibility with our pipeline.
An ablation study was performed using the GrAco dataset. For evaluation, we compared our results against the provided ground truth: high-precision GNSS data (for outdoor sections, with reported accuracy of ±1cm~\cite{fengS3EMultiRobotMultimodal2024,zhuGRACOMultimodalDataset2023}) and motion capture systems (for indoor sections, with millimeter-level precision~\cite{fengS3EMultiRobotMultimodal2024}), using the \textit{evo} package~\cite{gruppEvoPythonPackage2017}.
In our experiments, all techniques share the same odometry estimates to isolate the impact of map merging.

\subsection{Full System Performance}
\ExecuteMetaData[figures/experimentation_figures.tex]{s3e-table}
In~\cref{fig:viz-s3e-all}, we illustrate \gls{mold}’s performance on the challenging S3E Dormitory sequence, demonstrating a closer alignment with the ground truth compared to the Swarm-SLAM baseline.
The visualizations are supported by the detailed results presented in~\cref{tab:s3e}, where we report the \textit{Average Translation Error} (ATE), the number of loop closures $N$, and the computation time required by the factor graph solver for each technique on four sequences.

The best performance was achieved by \gls{mold} using the smoothed scales formulation (\textbf{SS}) optimized with the \textbf{LM} solver. This approach generated orders of magnitude more loop closures, significantly enhancing the accuracy of the resulting solution. To identify successful matches, we applied a correspondences ratio threshold $R_{thr}$ of $0.3$. While some outliers might have passed through, their effects were mitigated by the ratio-based confidence mechanism (see~\cref{eq:confidence}), without needing a more computationally expensive robust solver like GNC.
Even though a few accurate loop closures may suffice for map merging, our approach’s ability to generate more loop closures is a significant advantage, especially in scenarios where potential loop closures are scarce.
In addition, directly addressing the scale ambiguity—without relying on approximate scale estimates or costly outlier rejection mechanisms—significantly enhances performance.

The improved accuracy of \gls{mold} with smoothed scales (SS) indicates that loop closure scales within clusters may indeed be correlated. This makes the additional computation time versus IS—observed in cases like the Dormitory sequence—a worthwhile trade-off when resources allow.
Notably, on the S3E Dormitory sequence, the stereo Swarm-SLAM configuration was unable to merge all three trajectories due to the challenges of detecting loop closures from opposite viewpoints using traditional stereo matching techniques. Similarly, the LiDAR configuration of Swarm-SLAM performed poorly on this sequence due to significant perceptual aliasing in the structurally repetitive dormitory hallways.



To further analyze our proposed approach, the following subsections present an ablation study we conducted to evaluate the effectiveness and impact of the various novel components.

\subsection{Relative Pose Estimation Accuracy and Robustness}
\label{subsec:ablation-rel-pose}
In our first ablation experiment, we used the GrAco dataset~\cite{zhuGRACOMultimodalDataset2023}, which involves six robots, and extracted 88,096 image pairs that are less than 10 meters apart according to GNSS data. We then applied our registration pipeline to each pair to obtain the relative poses.

\ExecuteMetaData[figures/experimentation_figures.tex]{conf-pr}

The first issue we investigated was determining the validity of the matches. Due to perceptual aliasing, place recognition sometimes fails, incorrectly identifying different locations as the same. This challenge is exacerbated by DUSt3R~\cite{wangDUSt3RGeometric3D2024}, and subsequently MASt3R~\cite{leroyGroundingImageMatching2024}, which are now capable of matching images from almost any viewpoint with confidence, making it harder to verify which image matches correspond to valid relative poses and should be included as inter-robot loop closures in the pose graph. In~\cref{fig:conf-pr}, we evaluate the precision and recall of five different metrics. A match is considered an outlier if its translation error compared to the ground truth exceeds 2 meters, with the measurements scaled using ground truth data.
We found that relying solely on CosPlace similarity is insufficient to distinguish registration inliers from outliers. Alternatively, we considered metrics based on MASt3R's outputs. We considered the average confidence produced by MASt3R and computed the confidence ratio between the loop match and the odometry match. We also considered the number of feature correspondences and the correspondences ratio. The results show that both the number of correspondences and the correspondences ratio provided strong performance, with the ratio being preferable due to its unitless nature, making it easier to tune compared to the number of correspondences, which is affected by the image size and field of view.

\subsection{Multi-Robot Scale and Pose Graph Optimization}
\label{subsec:ablation-scale-pgo}
In the following experiments, we evaluate the full pose graph solution involving the full 6-robot GrAco sequence. We mapped the feature correspondences ratio to a confidence metric (see~\cref{eq:confidence}) to weight the inter-robot loop closures during pose graph optimization.

\ExecuteMetaData[figures/experimentation_figures.tex]{opt-vins-table}

\ExecuteMetaData[figures/experimentation_figures.tex]{opt-lio-table}
In~\cref{tab:graco-vins,tab:graco-lio}, we analyze the effects of different factor graph formulations, and the impact of the odometry backbone. We tested two odometry backbones: VINS-Mono~\cite{qinVINSMonoRobustVersatile2018}, making the system fully monocular, and the more accurate LiDAR-based LIO~\cite{shanLIOSAMTightlycoupledLidar2020}, to demonstrate our approach's effectiveness when the odometry has near-perfect scale. While monocular odometry yielded reasonable results, inaccuracies in scale at various parts of the trajectory—particularly at the start, where the online scale estimation had not yet converged—led to reduced performance. Since the GrAco dataset lacks VIO calibration sequences, all robot trajectories were affected by these initial scale inaccuracies. In real-world scenarios, proper calibration remains critical for effective VIO deployments. Results from our approach combined with the LIO-based solution demonstrate impressive accuracy, with errors below 4 meters over $3.5$ kilometers of trajectories.

For each backbone, we compared the formulations previously introduced. 
For completeness, we also report results for both the non-robust LM solver and the robust GNC solver for each of our factor graph formulations.
Additionally, we report the computation time on both our robot onboard computer, and a desktop server equipped with an AMD Ryzen 7 3700X CPU and an NVIDIA RTX 3070 GPU. This comparison is interesting given that some other existing C-SLAM approaches offload state estimation to servers~\cite{schmuckCOVINSVisualInertialSLAM2021}.
Note that the pose graph optimization runs on CPU, while both place recognition and registration are perfomed on GPU.
We provide each result with different scale initialization techniques, whether using ground truth, directly the raw output from MASt3R, or our odometry-based scaling (see~\cref{eq:scaleinit}).
The results in~\cref{tab:graco-vins,tab:graco-lio} confirm that our formulations with independent or smoothed scales achieve the best—or near-best—performance, without the need for the expensive robust optimization of GNC.
This shows that the use of a robust solver offers limited benefits. In fact, it can decrease performance when paired with the smoothed scale formulation, potentially because the outlier rejection mechanism of GNC conflicts with our confidence-based weighting.
Compared to the several minutes required by GNC, our optimization runs in under a second on the server and just a few seconds on the onboard computer.

\ExecuteMetaData[figures/experimentation_figures.tex]{split-compute}
\ExecuteMetaData[figures/experimentation_figures.tex]{sparse-kf-results}

\subsection{Resource Efficiency}

In addition to the computational efficiency provided by our scale-aware optimization compared to robust methods, our approach offers further efficiency advantages. Specifically, we split the registration into two stages where keyframes are encoded locally on each of the two robots involved in the loop closure, and decoding is done on only one of the robots, as outlined in~\cref{fig:mast3r}. This way, a keyframe part of multiple loop closures only needs to be encoded once—particularly useful when managing a large number of loop closure candidates. In~\cref{fig:split-compute},
we show the computation gain on the GrAco dataset against a naïve baseline where encoding of both images is done for all candidates.
We can see that MASt3R inference is quite costly on a robot onboard computer (in blue) compared to a server with a larger GPU (in orange).
This matches our observations for the pose graph optimization in~\cref{tab:graco-vins,tab:graco-lio}, where onboard computing is approximately 3 to 4 times slower than the server.
Thus, further gains could be achieved through more advanced load-sharing strategies between robots and servers.
Another resource efficiency gain comes from reducing the number of keyframes to process and store in memory. However, reducing keyframes typically leads to decreased accuracy as the mapping becomes more coarse and fewer loop closures can be detected. In~\cref{fig:sparse-kf}, we assess how increasing the distance between keyframes—and thus proportionally reducing their number—affects the ATE of the solution. We also compared this with loop closures computed using SuperPoint+SuperGlue and OpenCV.
While other methods show a rapid performance decline as the keyframe distance increases, our approach maintains low ATE values even with distances of up to 4 meters. This demonstrates its ability to support sparser and more memory-efficient C-SLAM solutions without sacrificing accuracy.
The efficiency gains of using sparser keyframes extend beyond memory savings, as sparse maps can also be compared—via place recognition and registration—using less communication bandwidth and in less time.
High-accuracy sparse maps are beneficial for deployments on small robots or consumer electronics.

\section{Conclusion}
\label{sec:conclusion}

In this paper, we introduced \gls{mold}, a novel decentralized collaborative SLAM approach that leverages 3D foundation models to address the challenges of multi-robot loop closure detection. By incorporating monocular pose estimation and scale optimization, \gls{mold} effectively improves inter-robot loop closure computation in scenarios where current methods struggle due to large differences in robot viewpoints.
Our experimental results, and ablation study, demonstrate that \gls{mold} outperforms state-of-the-art approaches, particularly in terms of accuracy. We showed that when accurate odometry with the correct scale is available, it is possible to easily and efficiently integrate up-to-scale loop closures in multi-robot pose graph optimization.
Our approach, however, relies on the availability of metric-scale odometry from the single-robot systems to anchor the scale of the global map. It also assumes the odometry is stable enough to provide a reliable baseline for our confidence metric.
Furthermore, while the pre-trained MASt3R model generalizes impressively, performance may degrade in domains drastically different from its training data (e.g., underwater imagery), which would necessitate fine-tuning.

Looking ahead, there is considerable potential for tighter integration between 3D foundation models and C-SLAM systems to improve measurement accuracy and the overall representation of explored environments.
Progress in this area would also be greatly accelerated by the creation of more diverse, large-scale C-SLAM datasets that feature varied sensor types and challenging, realistic multi-robot trajectories.

\bibliographystyle{IEEEtran}
\bibliography{small_ref}

\vfill\pagebreak

\end{document}